\pdfoutput=1

\documentclass[11pt]{article}

\usepackage[preprint]{acl}

\usepackage{times}
\usepackage{latexsym}
\usepackage{xcolor}  
\usepackage{xspace}

\usepackage[T1]{fontenc}

\usepackage[utf8]{inputenc}

\usepackage{microtype}

\usepackage{inconsolata}

\usepackage{graphicx}
\usepackage{amsfonts}
\usepackage{amsmath}
\usepackage{makecell} 
\usepackage{tabularray}
\usepackage[most]{tcolorbox}
\usepackage{enumitem}
\usepackage{multirow}
\usepackage{booktabs}
\usepackage{authblk}
\newcommand{\correspondauthor}{\textsuperscript{*}}
\newcommand{\modelname}{\textsc{RaPID}\xspace}

\newcommand{\eg}{\emph{e.g.,}\xspace}

\title{\modelname: Efficient Retrieval-Augmented Long Text Generation with \\ Writing Planning and Information Discovery}

\author{
 \textbf{Hongchao Gu\textsuperscript{1}},
 \textbf{Dexun Li\textsuperscript{2}},
 \textbf{Kuicai Dong\textsuperscript{2}},
 \textbf{Hao Zhang\textsuperscript{2}},
 \textbf{Hang Lv\textsuperscript{1}},
\\
 \textbf{Hao Wang\textsuperscript{1}}\correspondauthor,
 \textbf{Defu Lian\textsuperscript{1}},
 \textbf{Yong Liu\textsuperscript{2}},
 \textbf{Enhong Chen\textsuperscript{1}}\\
\textsuperscript{1} University of Science and Technology of China, \textsuperscript{2}Huawei Noah's Ark Lab
\\
\texttt{\{hcgu, lvhang1001\}@mail.ustc.edu.cn} 
\texttt{\{wanghao3, liandefu, cheneh\}@ustc.edu.cn} 
\\
\texttt{\{lidexun, dong.kuicai, zhang.hao3, liu.yong6\}@huawei.com}
}

\begin{document}
\maketitle
{
\renewcommand{\thefootnote}{*}
\footnotetext[1]{Corresponding authors.}
\renewcommand{\thefootnote}{\fnsymbol{footnote}}
}

\begin{abstract}
Generating knowledge-intensive and comprehensive long texts, such as encyclopedia articles, remains significant challenges for Large Language Models. It requires not only the precise integration of facts but also the maintenance of thematic coherence throughout the article.
Existing methods, such as direct generation and multi-agent discussion, often struggle with issues like hallucinations, topic incoherence, and significant latency.
To address these challenges, we propose~\modelname, 
an efficient retrieval-augmented long text generation framework.
\modelname~consists of three main modules: (1) Retrieval-augmented preliminary outline generation to reduce hallucinations, (2) Attribute-constrained search for efficient information discovery, (3) Plan-guided article generation for enhanced coherence. Extensive experiments on our newly compiled benchmark dataset, FreshWiki-2024, demonstrate that \modelname~significantly outperforms state-of-the-art methods across a wide range of evaluation metrics (\eg long-text generation, outline quality, latency, etc). Our work provides a robust and efficient solution to the challenges of automated long-text generation.

\end{abstract}
\section{Introduction}
\begin{figure}[h]
  \centering
  \includegraphics[width=1\linewidth]{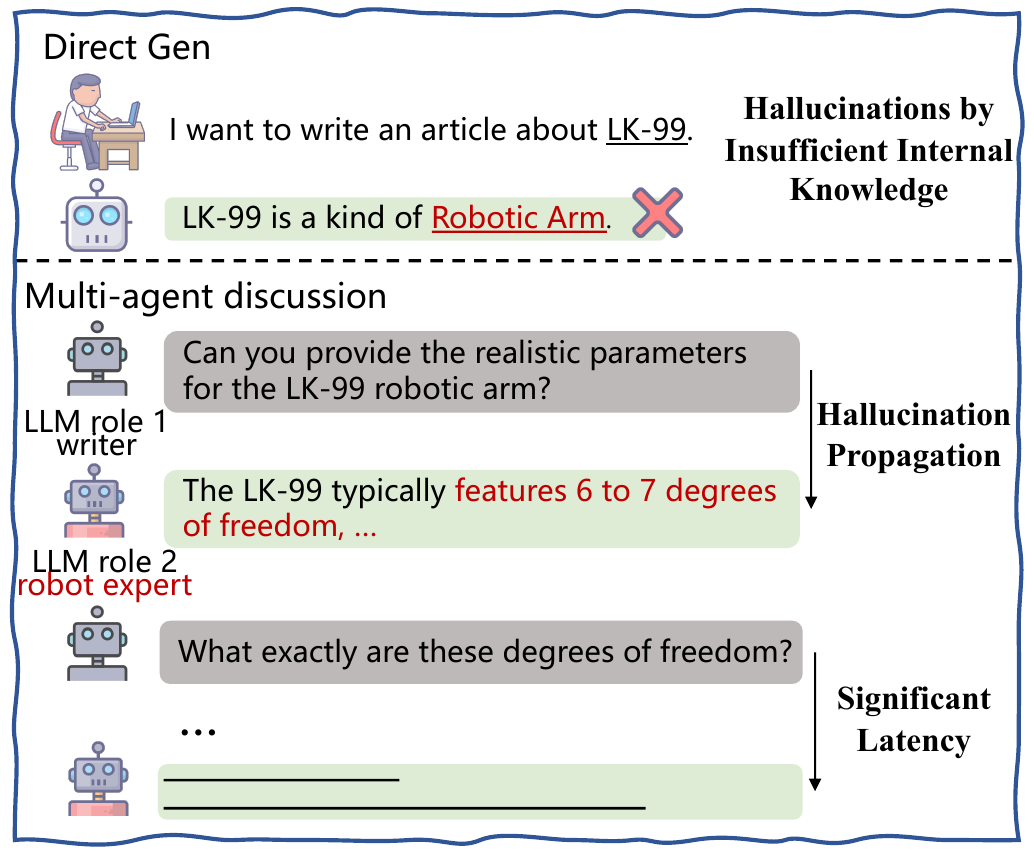}

  \caption{An example of generating a wiki-style article using various methods highlights distinct challenges and considerations. Direct generation may suffer from the large model's limited internal knowledge. While methods based on multi-agent discussions can provide broad coverage of the topic, they may also result in increased hallucinations and reduced efficiency.}

  \label{fig:wikistart}
\end{figure}

Large Language Models (LLMs) have showcased significant proficiency in handling various natural language tasks~\citep{chen-etal-2024-exploring-potential,hu-etal-2024-gentranslate,chen-etal-2024-semantic}, achieving near-human performance in tasks like summarizing lengthy documents~\citep{liu-etal-2024-learning} and crafting poetry~\citep{yu-etal-2024-charpoet}. 
Despite these achievements, generating long and knowledge-intensive texts (\eg encyclopedia articles) faces significant challenge~\citep{shen2023summarizationdesigningaisupport}. Such tasks demand not only the seamless integration of facts and narratives over extensive text but also the maintenance of thematic and stylistic consistency throughout the document.

\citet{bai_longwriter_2024} focus on fine-tuning LLMs to generate long-text solely based on LLMs' internal knowledge.
However, 
limited LLM's internal knowledge~\cite{you2023eipetextevaluationguidediterativeplan} can lead to significant hallucinations in its generation,
as evident in Figure~\ref{fig:wikistart}.
Such phenomenon is further exacerbated when generating with fact-intensive content~\cite{rawte-etal-2023-troubling}.
To address these issues, retrieval-augmented generation (RAG) techniques are commonly utilized to provide LLMs with additional retrieved content \citep{jiang2024longragenhancingretrievalaugmentedgeneration}. Nonetheless, the effectiveness and scope of retrieved content can significantly influence the quality of generated long texts \citep{shao-etal-2024-assisting}.
Moreover, RAG-based long text generation relies on multiple rounds of retrieval, formulating dedicated search queries, and obtaining characteristic content for different sections. This can cause thematic and stylistic inconsistency between sections~\cite{zhou_recurrentgpt_2023}.
A more effective approach is to mimic human writing practices~\citep{prewriting-stage, doyle1994information}, directing LLMs to develop a structured outline prior to full-text generation.

A recent work, STORM~\citep{shao-etal-2024-assisting} introduces a novel long-text generation strategy based on multi-agent discussions. This approach identifies multiple perspectives on each topic and uncovers information through interactions among agents assigned distinct roles, resulting in better performance compared to direct RAG-based methods.

Despite its promising results, long text generation based on multi-agent disscussion faces three challenges as follows:
\textbf{(1)} \textit{writing intentions are often general and ambiguous}
,  but agents are constrained by the internal knowledge of LLMs, making it highly possible to misinterpret the writing intention from the very beginning and generate hallucinations.
\textbf{(2)} \textit{agent interactions lack self-correction mechanisms}, leading to a failure to detect hallucinations and allowing ineffective discussions to perpetuate errors unchecked.
\textbf{(3)} \textit{long texts inherently involve intricate long-range dependencies and complex logical structures}. Consequently, maintaining consistency and accuracy throughout the generation process remains challenging.

To address these challenges, we propose \modelname, an efficient \textbf{R}etrieval-\textbf{A}ugmented long text generation framework with writing \textbf{P}lanning and \textbf{I}nformation \textbf{D}iscovery. Specifically,
\textbf{(1)} we design a retrieval-augmented outline generation module.
We first establish an outline corpus containing approximately 2.6 million example outlines from Wikipedia, covering a diverse range of topics and structures.
Given a question, it is disambiguated via results from web search, and the refined intent is used to retrieve outline examples from established outline corpus. These outline examples are used as context for high-quality outline generation.
\textbf{(2)} To comprehensively gather information across multiple topics, we extract attributes from the outline. We maintain a comprehensive attribute buffer and convert the extracted attributes into search queries for attribute-constrained search. The collected information contributes to iteratively updating the outline and serves as reference material to support the generation of the final long text.
\textbf{(3)} Most importantly, we design a novel structured writing plan. It is derived from the outline, to guide full article generation. 
Specifically, the writing plan is a topological graph that plots section dependencies and writing sequences.

Our contributions are summarized as follows:
\begin{itemize}[left=0pt, itemindent=0pt,itemsep=0pt]
    \item 
    We revisit the automated knowledge-intensive long text generation, particularly focusing on the unified consideration of pre-writing and generation stages for generating wiki-style articles.
    \item We propose \modelname, a framework that leverages retrieval-augmented outline generation, efficient information discovery, and logic writing plan guidance to generate comprehensive and knowledge-intensive articles.
    \item We construct a new dataset, and extensive experiments
    demonstrate the effectiveness of \modelname in terms of factuality and coherence.
\end{itemize}
\section{Related Works}
\paragraph{Long-form Text Generation}
Long-form text generation~\citep{tan-etal-2021-progressive,guan-etal-2021-long,yang-etal-2022-long} has been a significant focus in NLP research, even prior to the emergence of large language models. Recently, researchers have achieved considerable success in various applications of long text generation using LLMs, including creative writing~\citep{lei_ex3_2024,li-etal-2024-advancing,yang-etal-2023-doc}, scientific survey~\citep{kang_researcharena_2024,wang2024autosurveylargelanguagemodels} and expository writing~\citep{balepur-etal-2023-expository}. Previous efforts to enhance long text generation have often involved improving models' abilities to produce extended outputs by constructing crafted SFT dataset~\cite{xu2024chatqa2bridginggap}. This paper, however, primarily focuses on generating knowledge-intensive long articles based on topic retrieval. For instance, ~\citet{shao-etal-2024-assisting} proposed a pre-writing method that utilizes perspective-guided questioning and outline generation, which automates wiki-style article generation from scratch. Building on this foundation,~\citet{jiang2024unknownunknownsengagedhuman} introduced a multi-agent dialogue and user interaction mechanism, which further enhances the large language model's ability to explore unknown unknowns. Additionally,~\citet{pham2024surimulticonstraintinstructionfollowing} and~\citet{bai_longwriter_2024} proposed to enhance the ability of LLMs to directly generate long outputs by improving LLM alignment methods. However, these methods still face challenges related to hallucinations, efficiency, and consistency, which are the primary focus of optimization in this paper.

\paragraph{Retrieval Augmented Generation}
Retrieval-Augmented Generation~\cite{lewis2020retrieval} represents an innovative approach that integrates the advantages of information retrieval with language generation. The performance of RAG heavily depends on the accuracy of the retrieved information~\citep{mallen-etal-2023-trust}. Some works improve retrieval performance through adaptive retrieval~\citep{asai2023selfrag,liu2024ctrlaadaptiveretrievalaugmentedgeneration} or query rewriting~\citep{gao-etal-2023-precise,ma-etal-2023-query}. 
More recently,~\citet{edge2024localglobalgraphrag} proposes to enhance the accuracy of RAG by automatically building knowledge graphs.
Although these methods have demonstrated significant effectiveness in short-format output tasks~\citep{joshi-etal-2017-triviaqa,yang2018hotpotqa}, they remain inadequate for application in knowledge-intensive long text generation tasks, which usually require multi-faceted information retrieval and fine-grained information filtering.


\label{section:methods}
\begin{figure*}[h]
  \centering
  \includegraphics[width=1\linewidth]{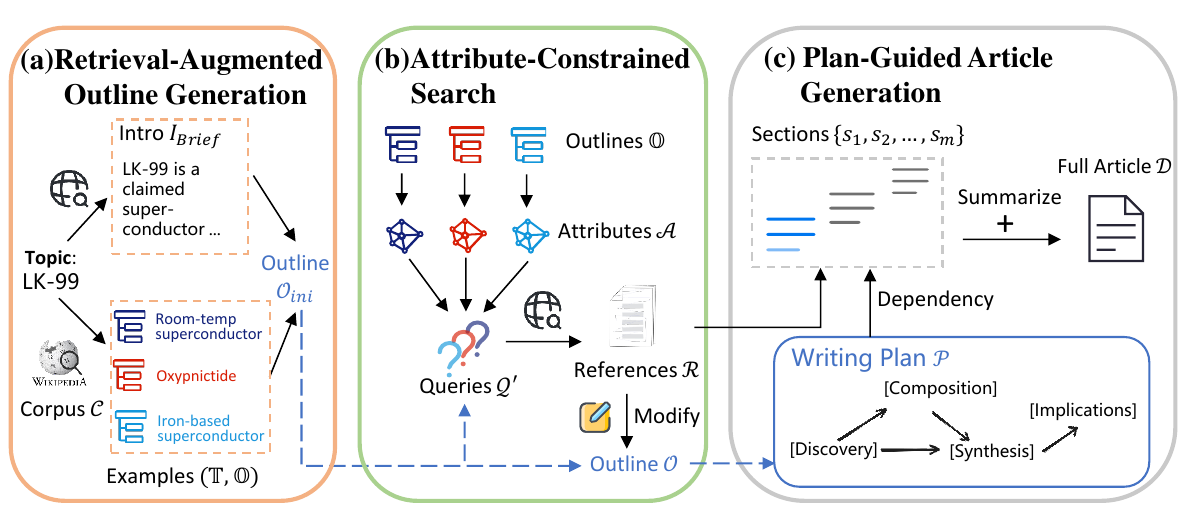}
    
\caption{The framework of \modelname, which consists of three main stages: (a) Retrieval-Augmented Outline Generation, where an initial outline is created based on a brief introduction and examples; (b) Attribute-Constrained Search, which leverages an attribute-based mechanism to discover relevant information and refine the outline accordingly; and (c) Plan-Guided Article Generation, where a structured writing plan is developed based on dependencies between sections, resulting in a more coherent and fluent article. The blue dashed lines illustrate how the outline evolves throughout the processes of information discovery and writing planning.}

  \label{fig:main}
\end{figure*}

\section{Methods}

In this section, we introduce \modelname, an efficient and effective framework for automated wiki-style article generation. We first formulate the article generation problem in Section~\ref{sec:definition}. Next, we detail the retrieval-augmented outline generation process in Section~\ref{sec:outline_gen}, the attribute-constrained search in Section~\ref{sec:info_seek}, and plan-guided long-text generation in Section~\ref{sec:article_gen}. The overall framework of \modelname is illustrated in Figure~\ref{fig:main}. 

\subsection{Problem Formulation}
\label{sec:definition}
The goal of this task is to automatically generate a comprehensive wiki-style article, denoted as $\mathcal{D}$, based on a given topic $t$. This process involves generating a clear outline and crafting coherent sections that thoroughly explore the topic. Formally, the task includes three main steps: (i) Initial outline generation: given a topic $t$, we first generate an initial outline $\mathcal{O}_{ini}$ related to $t$, (ii) Information gathering and update: we then use the search engine to collect a diverse set of information from the knowledge source $\mathcal{S}$ as the reference $\mathcal{R}$ and iteratively refine $\mathcal{O}_{ini}$ to produce the final outline $\mathcal{O}$; (iii) Article crafting: finally, the complete article $\mathcal{D}$ is generated by elaborating on the final outline $\mathcal{O}$ using the reference $\mathcal{R}$.

\subsection{Retrieval Augmented Outline Generation}
\label{sec:outline_gen}

\emph{``Well begun is half done.''} A clear and logically coherent outline is crucial for the writing process, which serves as a roadmap to guide the detailed content unfold in appropriate places and to ensure overall cohesion~\citep{fan-gardent-2022-generating}.
However, existing methods rely on generating an initial outline using the model's internal knowledge, which will inevitably introduce inaccuracies or misconceptions. This outline is then iteratively refined through RAG. However, such inaccuracies may escalate as the writing progresses, leading to topic deviations. To address this, we employ retrieval-augmented generation during the initial outline generation phase to improve writing intent recognition and reduce hallucinations. Additionally, we retrieve similar topic outline examples from a curated corpus, generating a high-quality preliminary outline that is further refined through an efficient information retrieval module.

\paragraph{Intent recognition.}
To reduce ambiguity, we begin by using an intention recognition module to clarify intentions and maintain focus. Specifically, for a given topic $t$, we first query a search engine for relevant information and then generate an initial summary of the results. This process is formalized as follows:
\begin{equation}
\mathcal{S}_t= \texttt{Sch}(t,\mathcal{S})   \text{ and } I_{\text{brief}} = \texttt{LLM}(t, \mathcal{S}_t)
\end{equation}
where $I_{\text{brief}}$ represents the brief summary of the topic, and $\mathcal{S}_t$ denotes the search results obtained by the search engine $\texttt{Sch}$ from the knowledge base $\mathcal{S}$. 

\paragraph{Outline example retrieval.}
To improve the quality of the generated initial outline, we retrieve high-quality outline examples from a curated corpus based on the given topic, which serve as few-shot examples for outline generation. Specifically, to construct the corpus $\mathcal{C}$, we extract outlines from the extensively crawled structured articles in the Wiki archives\footnote{\url{https://dumps.wikimedia.org/}}. We then concatenate the topic $t$ and its brief summary $I_{\text{brief}}$ to facilitate query expansion, and use a dense retriever $\texttt{Ret}_1$ to retrieve similar topics from the constructed Wiki corpus $\mathcal{C}$. We denote these retrieved topics as $\mathbb{T}=\{t'_1,t'_2,...,t'_n\}$, with their corresponding outlines $\mathbb{O}=\{\mathcal{O}'_1,\mathcal{O}'_2,...,\mathcal{O}'_n\}$. We have:
\begin{equation}
    (\mathbb{T}, \mathbb{O}) = \texttt{Ret}_1 (t, I_{\text{brief}}, \mathcal{C})
\end{equation}

After retrieval, we combine the topic $t$, summary $I_{\text{brief}}$ as the reference context, and the retrieved outlines $\mathbb{T}$ and $\mathbb{O}$ as few-shot examples to instruct LLM to generate the initial outline, denoted as $\mathcal{O}_{ini}$:
\begin{equation}
    \mathcal{O}_{ini} = \texttt{LLM}(t, \mathcal{S}_t, \mathbb{T}, \mathbb{O})
\end{equation}

\subsection{Attribute-Constrained Search}
\label{sec:info_seek}
After generating the initial outline $\mathcal{O}_{ini}$, we design an attribute-constrained information collection module to enhance the efficiency and comprehensiveness of information gathering, further refining the $\mathcal{O}_{ini}$ to produce the final outline $\mathcal{O}$.

\paragraph{Attribute \& Query Generation.}
In the process of generating long-form text, it is crucial to ensure that the content is comprehensive and relevant. To this end, we introduce the concept of attributes. Each attribute is defined as a distinct, indivisible concept to prevent overly complex queries, thereby effectively summarizing the essential information for writing the full article. By breaking down the outline into distinct attributes, we can focus on capturing the most pertinent details without introducing overwhelming complexity. We prompt LLM to extract the possible attribute from the current outline $\mathcal{O}_{ini}$ and maintain an attribute buffer, $\mathcal{A}$. Such buffer encompasses the key aspects necessary for composing the final article. We then prompt LLM to transform the attributes into queries, $\mathcal{Q}$, that are related to the article's title and are suitable for search engine retrieval~\footnote{Note that such attributes can also be directly provided or modified by humans to guide relevant retrieval during long-form text generation, incorporating points that are of interest or that they wish to highlight.}. The collected information will be added into the reference set $\mathcal{R}$. We have:
\begin{equation}
    \mathcal{A} = \texttt{LLM}(\mathcal{O}_{ini}), \mathcal{Q}=\texttt{LLM}(\mathcal{A})\text{ and } \mathcal{R}=\texttt{Sch}(\mathcal{Q},\mathcal{S})
\end{equation}

We have fully parallelized the search process to enhance the efficiency.
The example queries are then selectively merged into a set of queries $\mathcal{Q}$ by a LLM. The merged queries $\mathcal{Q}'$ are used in parallel with search engines to gather information, which serves as references $\mathcal{R}$ for the outline refinement and subsequent writing process. This attribute-constrained approach is highly parallelized, enabling effective and efficient information gathering.

\paragraph{Outline Refinement}
To further align the initial outline $\mathcal{O}_{ini}$ with the desired structure, we provide the LLMs with the titles of all references from $\mathcal{R}$ and instruct LLMs to generate modifications to $\mathcal{O}_{ini}$. These modifications can include actions such as $\texttt{[add section]}$, $\texttt{[delete section]}$, or $\texttt{[do nothing]}$, ensuring that the outline aligns with the most current information and insights available.

We will iterate the outline refinement process until the attribute buffer $\mathcal{A}$ and outline converge, or the search limits are reached. Thus, the final outline $\mathcal{O}$ is generated based on $\mathcal{O}_{ini}$ and $\mathcal{R}$:
\begin{equation}
    \mathcal{O} = \texttt{LLM}(\mathcal{O}_{ini}, \mathcal{R})
\end{equation}

\subsection{Plan-Guided Article Generation}
\label{sec:article_gen}
\paragraph{Plan Generation}
Existing methods directly generate the article based on outline $\mathcal{O}$, such a simple one-way writing logic may cause the content to gradually deviate from the initial section. A long article has a relatively complex structure, with interdependencies between sections. To maintain the overall coherence and consistency, we introduce an additional writing plan, denoted as $\mathcal{P}$, which provides a carefully planned writing order derived from the outline $\mathcal{O}$. Specifically, we first extract all the first-level section titles from $\mathcal{O}$, denoted as $\{s_1, s_2, \dots, s_m\}$. We then define the ``dependency'' of a section as the content from other sections that must be referenced before developing that particular section, ensuring a logical flow and integration throughout the article.

Leveraging the planning capabilities of LLMs, we construct plan  $\mathcal{P}$ by providing human-written few-shot examples. Specifically, the plan $\mathcal{P}$ consists of a list that defines the dependencies of each section, with ``None" indicating no dependencies on other sections. We then calculate its topological order to ensure the feasibility of the plan, and the full article is generated section by section according to this order. In the rare instances where the generation of $\mathcal{P}$ fails, we will revert to the more conventional approach of generating each section in parallel.

The generation of the particular section $s_i$ can be formulated as:
\begin{align}
    \mathcal{P} &= \texttt{LLM}(\{s_1, s_2, \dots, s_m\}), \nonumber \\ 
    \mathcal{R}_{s_i}&= \texttt{Ret}_1(s_i, \mathcal{P}, \mathcal{R},\mathcal{O}),  \\
    s_i &= \texttt{LLM}(t, \mathcal{P}, \mathcal{O}, \mathcal{R}_{s_i}). \nonumber
\end{align}

where $\mathcal{R}_{s_i}$ represents the references retrieved from $\mathcal{R}$ that are needed to generate $s_i$.

After generating all the sections, we follow established methodologies by concatenating them to create a comprehensive summary of the topic. This process culminates in the creation of a complete wiki-style article.

\section{Experiments}
\subsection{Dataset}
To evaluate the capabilities of \modelname, we introduce the FreshWiki-2024 dataset, a chronological extension of the original FreshWiki dataset~\citep{shao-etal-2024-assisting} which consists of human-authored Wikipedia articles covering 100 distinct topics. To address temporal data contamination concerns, FreshWiki-2024 follows the rigorous curation protocol of its predecessor and only includes Wikipedia entries revised during 2024, ensuring all evaluation materials postdate conventional model training cutoffs. For robust performance evaluation, we randomly selected 100 representative topics from FreshWiki-2024 to evaluate both \modelname and baseline methods. Comprehensive dataset statistics and details are provided in Appendix~\ref{app:dataset_stats}.

    \setlength{\tabcolsep}{3.5pt}

\begin{table*}[t]
\centering
\resizebox{\textwidth}{!}{
\small
\begin{tabular}{cl|c@{\hskip 3pt}c@{\hskip 3pt}c@{\hskip 3pt}c@{\hskip 3pt}|ccc|ccc|c}

\toprule
 \multicolumn{2}{c}{\multirow{3}{*}{\textbf{Method}}} & \multicolumn{4}{|c}{\textbf{LLM Evaluation}} & \multicolumn{3}{|c}{\textbf{Similarity}} & \multicolumn{3}{|c|}{\textbf{Factuality }} & \textbf{Info} \\
 \cmidrule(lr){3-6}
 \cmidrule(lr){7-9}
 \cmidrule(lr){10-12}
 
 & & Intere. & Organiz. & Relev. &  Cover. & Rouge-1 & Rouge-L & Entity Recall &  Precision & Claims & F1@300 & \textbf{Diversity}\\

\midrule 

\parbox[t]{3.4mm}{\multirow{5}{*}{\rotatebox[origin=c]{90}{\textit{\textcolor{gray}{\textbf{Qwen-Max}}}}} } 

 & RAG        & 3.66 & 4.24 & 4.56 & \textbf{4.20} & 33.22 & 14.72 & 11.23 & 46.50 & 193.90 & 54.09 & 0.456 \\
  & oRAG       & 3.32 & 3.79 & 3.87 & 3.61 &  40.04 & 14.49 & 9.27 & 43.90 & 396.30 & 65.90 & 0.552 \\
  & STORM      & 3.90 & 4.44 & 4.49 & 4.15 & 41.35 & 15.07 & 10.85 & 44.30 & 440.90 & 68.08 & 0.618 \\
  & Co-STORM   & 3.92 & 4.16 & 4.51 & 4.10 & 32.43 & 13.08 & 8.38 & 49.00 & 230.90 & 59.88 & 0.580 \\
  & Rapid & \textbf{3.96} & \textbf{4.49} & \textbf{4.69} & 4.13 & \textbf{44.65} & \textbf{15.86} & \textbf{12.60} & \textbf{51.10} & \textbf{493.70} & \textbf{77.98} & \textbf{0.650} \\
  \midrule 

\parbox[t]{3.4mm}{\multirow{5}{*}{\rotatebox[origin=c]{90}{\textit{\textcolor{gray}{\textbf{DeepSeek-v3}}}}} } 

 & RAG        & 4.03 & 4.39 & 4.66 & 4.05 & 25.27 & 12.23 & 9.40 & \textbf{57.50} & 103.10 & 43.02 & 0.469 \\
  & oRAG       & 3.68 & 4.23 & 4.70 & 4.11 & 24.74 & 10.74 & 6.53 & 52.40 & 125.40 & 46.50 & 0.498 \\
  & STORM      & 4.23 & \textbf{4.58} & 4.80 & 4.14 & 40.05 & 14.78 & \textbf{11.81} & 48.00 & \textbf{395.50} & 70.38 & 0.607 \\
  & Co-STORM   & 3.88 & 3.89 & 4.10 & 3.81 & 23.09 & 9.68 & 5.87 & 52.10 & 174.50 & 54.97 & 0.591 \\
  & Rapid & \textbf{4.37} & 4.53 & \textbf{4.85} & \textbf{4.21} & \textbf{40.64} & \textbf{15.16} & 10.94 & 54.70 & 337.70 & \textbf{73.62} & \textbf{0.670} \\
  \midrule 

\parbox[t]{3.4mm}{\multirow{6}{*}{\rotatebox[origin=c]{90}{\textit{\textcolor{gray}{\textbf{GPT-4o}}}}} } 

 & RAG        & 4.18 & 4.42 & \textbf{4.86} & 4.02 & 30.02 & 12.98 & 10.80 & 56.80 & 128.60 & 48.86 & 0.429 \\
  & oRAG       & 3.46 & 4.19 & 4.50 & 4.07 & 31.01 & 12.18 & 7.85 & 46.20 & 261.00 & 60.35 & 0.539 \\
  & STORM      & 4.16 & 4.46 & 4.61 & 4.09 & 40.56 & 14.46 & \textbf{12.19} & 45.60 & 398.10 & 67.88 & 0.626 \\
  & Co-STORM   & 4.31 & 4.30 & 4.78 & 4.14 & 34.14 & 13.07 & 8.83 & 44.80 & 276.80 & 60.31 & 0.584 \\
  & Rapid & \textbf{4.38} & \textbf{4.61} & 4.73 & \textbf{4.20} & \textbf{43.35} & \textbf{15.33} & 12.01 & \textbf{48.80} & \textbf{448.10} & \textbf{73.57} & \textbf{0.650} \\
  & ~ \textit{w/o plan} & 4.35 & 4.47 & 4.73 & 4.16 & 39.19 & 14.35 & 10.65 & 45.00 & 364.70 & 65.69 & 0.650 \\
  \bottomrule
\end{tabular}
}

\caption{
    Results of article quality evaluation. The best results of each metric are marked in \textbf{bold}. \textbf{LLM Evaluation} uses an LLM to assess the generated articles across four different dimensions. The LLM Evaluation uses a 1-5 scale. \textbf{Similarity} assesses the resemblance between generated and real articles. \textbf{Factuality} evaluates the quantity and accuracy of facts in the generated content. \textbf{Info Diversity} evaluates the diversity of the searched information.
}
\label{article_result}
\end{table*}

\subsection{Baselines}
We compared \modelname with several different LLM-based methods:
\begin{itemize}[leftmargin=*, itemsep=-0.4em, topsep=-0.0em]
    \item \textbf{RAG}: It generates an outline or full article based on the information retrieved from a search related to the writing topic.
    \item \textbf{Outline-driven RAG (oRAG)}: Building upon RAG, this method first generates an outline. It then uses the section titles to retrieve additional relevant information and generates the full article, section by section, using the retrieved content.
    \item \textbf{STORM}~\citep{shao-etal-2024-assisting}: A writing system designed to generate wiki-style articles from scratch. It majors in the pre-writing stage by generating the outline through retrieval and perspective-guided question-asking.
    \item \textbf{Co-STORM}~\citep{jiang2024unknownunknownsengagedhuman}: An enhanced version of STORM that incorporates user interaction and multi-agent discussion, further improving the exploration of previously unknown aspects within the topic.
\end{itemize}

\subsection{Evaluation Metrics}
\paragraph{Outline Quality.}
To objectively assess outline generation quality, we propose an evaluation framework based on section title alignment. Rather than employing NER-based metrics as in STORM (refer to Appendix~\ref{limitations_NER} for its limitations), we perform direct string matching between generated section titles and human-authored ground truth titles. We calculate recall, precision, and F1 score, providing a balanced assessment of outline comprehensiveness and conciseness.

\paragraph{Article Quality.}
Evaluating long-form text generation remains a challenging task. To comprehensively assess the quality of generated articles, we consider four key aspects:
(1) \textbf{Similarity}: We use Rouge~\cite{lin-2004-rouge} scores and entity recall from a FLAIR NER model~\cite{akbik2019flair} to compare the similarity between generated articles and human-written articles.
(2) \textbf{Factuality}: We also assess the factual accuracy of the generated articles. Specifically, we employ FActScore~\citep{min-etal-2023-factscore}, an automated long-text factuality evaluation framework based on LLMs. We report the precision and the number of claims of FActScore. Additionally, we include F1@300 as defined in FactAlign~\citep{huang2024factalignlongformfactualityalignment}, to penalize articles that contain few factual statements. 
(3) \textbf{LLM Evaluation}: To evaluate the quality of generated articles at the article level, we utilized \texttt{Prometheus-7b-v2.0}~\citep{kim2024prometheus}, an open-source model that supports a 5-point rubric for text evaluation. We employed the same version of the criteria used by STORM, focusing on four aspects: \textit{Interest}, \textit{Organization}, \textit{Relevance}, and \textit{Coverage}.
(4) \textbf{Diversity}: We measure the diversity of collected information using the \textit{info diversity} metric proposed by CoSTORM, which quantifies the variety of information included in the generated articles. Further details on the evaluation metrics can be found in Appendix~\ref{article_evaluation}.

\subsection{Implementation Details}
We implemented \modelname using the DSPy framework\footnote{\url{https://github.com/stanfordnlp/dspy}}, with all prompts detailed in Appendix~\ref{sec:prompt}. Three foundation models were employed as backbones: \texttt{gpt-4o-2024-11-20}, \texttt{qwen-max-2024-09-19}, and \texttt{deepseek-v3}. For web search functionality across all methods, we integrated the Google Custom Search API\footnote{\url{https://developers.google.com/custom-search/v1/overview}}, retrieving the top-5 most relevant results per query to balance precision and computational overhead. To prevent data leakage, we explicitly excluded official Wikipedia pages related to target topics during web search and outline retrieval. We implemented $\texttt{Ret}_1$ and $\texttt{Ret}_2$ using \texttt{e5-large-v2}~\citep{wang2022text} and \texttt{paraphrase-MiniLM-L6-v2}~\citep{reimers-2019-sentence-bert}, respectively. The Wikipedia dump\footnote{\url{https://dumps.wikimedia.org/backup-index.html}} (version 2024-08-01) served as the retrieval corpus for outline generation. We have made both the original Wikipedia dump and the FreshWiki-2024 dataset publicly available \footnote{\url{https://drive.google.com/drive/folders/1GNWE0ZEPijFpdjuPOfWLjWwEcQYq3Kz2?usp=sharing}}.

\section{Results and Analysis}

\subsection{Main Results}

Table~\ref{article_result} presents the article quality evaluation of \modelname across different foundation models. Our approach demonstrates substantial improvements over baseline methods in multiple dimensions. In terms of LLM-based evaluation and similarity metrics, \modelname outperforms existing baselines on most criteria while remaining competitive on the remaining indicators. The factuality assessment reveals that \modelname achieves significant improvements in F1@300 scores across all foundation models compared to baselines, indicating its superior capability to maintain factual accuracy while preserving high information density. Furthermore, \modelname attains the highest info diversity scores among all tested configurations, suggesting that our method effectively balances information discovery efficiency with diversity preservation. 

Additionally, we observed that certain simple baseline methods achieved unexpectedly high scores on specific metrics. For instance, RAG exhibits an unusually high relevance score on GPT-4o. This phenomenon may be attributed to the lower information density of its generated content, as evidenced by the fact that the number of claims it produces is less than one-third of that generated by \modelname. This observation underscores the challenge of evaluating long-form content quality using a single metric. In contrast, \modelname demonstrates superior performance across multiple evaluation criteria, reinforcing its robustness and effectiveness.

\begin{table}[t]
\centering
\small
\begin{tabular}{ll|ccc}
\toprule
  \multicolumn{2}{c|}{\textbf{Method}} & \textbf{Recall} & \textbf{Precision} &\textbf{F$_1$}    \\ 
\midrule
\parbox[t]{3.0mm}{\multirow{4}{*}{\rotatebox[origin=c]{90}{\textit{\textcolor{gray}{\textbf{QwenMax}}}}} } 

& RAG/oRAG   & 10.53 & 9.05    & 5.48  \\
 & STORM     & 13.86 & 6.23    & 4.94  \\
& Co-STORM   & 3.8 & 8.68    & 4.93  \\
& \textbf{ \modelname}      & \textbf{19.83} & 8.50    & \textbf{10.86} \\
\midrule

\parbox[t]{3.0mm}{\multirow{4}{*}{\rotatebox[origin=c]{90}{\textit{\textcolor{gray}{\textbf{DS-v3}}}}} } 

 & RAG/oRAG   & 9.97 & 5.79    & 6.82  \\
 & STORM     & 11.95 & 4.90    & 6.56  \\
& Co-STORM   & 5.13 & 8.78    & 5.82  \\
& \textbf{ \modelname}      & \textbf{20.82} & \textbf{14.25}    & \textbf{16.07} \\
\midrule

\parbox[t]{3.0mm}{\multirow{4}{*}{\rotatebox[origin=c]{90}{\textit{\textcolor{gray}{\textbf{GPT-4o}}}}} } 
 & RAG/oRAG   & 10.28 & 5.10    & 6.28  \\
 & STORM     & 11.12 & 3.66    & 5.19  \\
& Co-STORM   & 2.83 & 4.09    & 2.93  \\
& \textbf{ \modelname}      & \textbf{22.77} & \textbf{15.92}    & \textbf{17.52} \\
\bottomrule
\end{tabular}

\caption{
    Results of outline quality evaluation. We assess the resemblance between generated and real outline.  
}
\label{outline_result}
\end{table}

The quality evaluation results of outlines are presented in Table~\ref{outline_result}. \modelname significantly outperforms other methods across different foundation models, particularly in recall and F1-score metrics. This demonstrates its superior ability to maintain thematic focus while achieving comprehensive coverage of research topics. Notably, although STORM achieves relatively higher recall compared to other baselines, \modelname can efficiently generate higher-quality outlines through retrieval of similar topics, without requiring complex multi-turn agent discussion mechanisms.

\subsection{Ablation Studies}
To evaluate the impact of the writing plan, we conducted an ablation study on GPT-4o by removing the Plan-Guided Article Generation module and instead employing a parallel section-wise generation approach (denoted as "w/o plan"). As shown in Table~\ref{article_result}, the absence of a writing plan led to performance degradation across multiple metrics, particularly in organization and factuality. Without contextual paragraph awareness, the model exhibited stylistic inconsistencies and coherence issues between sections, explaining the decline in organization scores. Additionally, generating sections in isolation resulted in redundant content, as each paragraph attempted to cover multiple aspects of the topic independently. While techniques such as STORM can partially mitigate these issues through post hoc refinement, maintaining coherence and quality in long-form text remains challenging due to inherent output length limitations. In contrast, our structured planning approach explicitly defines inter-sectional dependencies, enhancing overall coherence and enabling the generation of longer, more cohesive articles. These findings underscore the critical role of planning in long-form text generation.

\begin{table}[t]
\centering
\small
\begin{tabular}{l|cccc}
\toprule
\textbf{Backbone} & \textbf{\makecell{Avg.\\Nodes}} & \textbf{\makecell{Avg.\\Edges}} & \textbf{\makecell{Dependency\\ Density}} & \textbf{\makecell{Longest\\Path}} \\ 
\midrule
\textbf{Qwen-Max}   & 8.72 & 16.64 & 2.09 & 2.79 \\
\midrule
\textbf{DeepSeek-v3}  & 7.21 & 12.18 & 1.91 & 2.62 \\
\midrule
\textbf{GPT-4o} & 8.43 & 11.86 & 1.55 &  3.59 \\
\bottomrule
\end{tabular}
\caption{
   Results of the graph metrics of Writing Plan.
}
\label{plan_analysis}
\end{table}

\subsection{Writing Plan Analysis}
Our method relies on constructing a directed acyclic graph (DAG) of sections prior to writing, which defines the contextual dependencies between sections. In this section, we analyze the Writing Plan's functionality through graph-theoretical metrics. By representing sections as nodes and inter-sectional dependencies as directed edges, we formally define the Writing Plan as a DAG structure $ G = (V, E) $, where $ V = \{v_1, v_2, \dots, v_n\} $ denotes the set of $ n $ nodes (sections), and $ E \subseteq V \times V $ represents the directed edges indicating dependencies. To quantify the "dependency density" of the Writing Plan, we define it as the ratio of the number of edges to the maximum possible edges in a minimally connected DAG:  
\begin{equation}
    D(G) = \frac{|E|}{n - 1},
\end{equation}
where $|E|$ is the cardinality of the edge set $ E $, and $ n - 1 $ corresponds to the maximum number of edges in a linear chain structure. For example, in a sliding window of length $ k = 1 $, where each section depends only on its immediate predecessor, the number of edges is $ |E| = n - 1 $, yielding a dependency density of $ D(G) = 1 $. Besides dependency density, we also report key metrics to describe the DAG structure: average nodes, edges, and longest path length. The longest path highlights the critical dependency sequence, determining the minimum steps to traverse the Writing Plan.

The results for the plans generated by three backbone models are summarized in Table~\ref{plan_analysis}. These results reveal that while the average number of sections across the plans remains relatively consistent, the number of dependencies exhibits more significant variation. Notably, the dependency density for all three backbones exceeds 1, ranging from 1.55 to 2.09. This indicates that the generated plans effectively transcend simple linear structures, capturing more complex long-range dependencies. These findings demonstrate the strong generalization capability of our method. The detailed case study of the writing plan is provided in Appendix~\ref{case_study}

Furthermore, an excessively long critical path can lead to reduced parallelism during section generation, thereby decreasing efficiency. However, the average longest path length of the plans generated by the three backbones does not exceed 4, ensuring that the generation time of \modelname remains within an acceptable range. Further evidence can be found in the detailed efficiency analysis provided in Appendix~\ref{Efficiency_Analysis}.

\section{Human Evaluation}
To gain deeper insights into the quality of the generated articles, we recruited master-degree level 10 volunteers to conduct a human evaluation. We randomly selected 20 topics and requested the volunteers to evaluate articles generated by both \modelname~and STORM for each topic. Each article was assigned to two different volunteers to ensure a balanced and unbiased assessment. We provided the volunteers with the same evaluation criteria as those used in the LLM-based assessment (see Table~\ref{table:llm_rubric}), which assess four dimensions: Interest Level, Organization, Relevance, and Coverage. Volunteers rated each dimension on a scale from 1 to 5. Additionally, they were asked to indicate their preference for each pair of articles. To mitigate potential bias, the order of methods in each article pair was randomized.

\begin{table}[t]
\centering
\small
\begin{tabular}{l|ccccc}
\toprule
Method & Intere. & Organiz. & Relev. & Cover. & \#Prefer \\
\midrule
STORM &  \textbf{4.01}  & 3.80 & 3.75 & 4.05 & 13      \\
\textbf{\modelname} & 3.96 & \textbf{4.03} & \textbf{4.25} & \textbf{4.15} & \textbf{27} \\
\bottomrule
\end{tabular}

\caption{Results of human evaluation. We had human evaluators evaluate the articles across four dimensions using a fine-grained scale ranging from 1 to 5, and we documented their preferences accordingly.}
\label{table:human_eval_result}
\end{table}

As shown in Table 3, our \modelname significantly outperforms STORM in terms of Organization, Relevance, and Coverage. Additionally, it achieves a notable improvement in overall preference, despite showing a slight disadvantage in Interest. These results indicate that \modelname generates articles of higher overall quality compared to STORM. 

Notably, the significantly lower Relevance score of STORM may be attributed to hallucination propagation caused by its agent-based discussion mechanism. For example, feedback from volunteers revealed that when processing the topic Vultures 1 (album), STORM’s pre-writing phase—relying solely on LLM-driven dialogue for information collection—misinterpreted the task, mistakenly assuming it involved generating content about vultures (the animal). This misunderstanding led to irrelevant, animal-related perspectives, rendering the entire information collection and writing process ineffective. In contrast, \modelname addresses this issue by employing retrieval-based methods to ensure accurate intent recognition from the outset. This approach enables a more precise and efficient information collection process, significantly reducing the risk of misinterpretation and ultimately improving the quality of the generated content.

\section{Conclusion}
In this paper, we addressed the challenge of knowledge-intensive long-text generation, with a particular focus on encyclopedia-style article generation. We introduced \modelname, an efficient and effective retrieval-augmented writing system that integrates a structured writing plan and information discovery mechanisms.
Our extensive experiments and human evaluations demonstrate that our model outperforms state-of-the-art baselines in both effectiveness and efficiency. Additionally, our model maintains scalability for other knowledge-intensive long-text writing tasks. In the future, we plan to explore the potential of LLMs in other domain-specific writing tasks, such as financial reports.

\section{Limitations}
Our proposed \modelname, while demonstrating efficacy in long-form Wikipedia article generation, is subject to two key constraints that warrant discussion. The current methodology exclusively focuses on textual content generation, overlooking the critical integration of multimodal elements such as images and tabular data that significantly enhance article credibility and reader engagement in authentic Wikipedia entries. Furthermore, the evaluation framework remains confined to the Wikipedia domain, leaving untested the method's adaptability to other long-form generation scenarios like financial reporting or academic writing, which demand distinct structural precision and factual rigor. These limitations underscore the need for subsequent research to expand multimodal integration capabilities and conduct cross-domain validation, thereby enhancing both the practical utility and generalizability of the proposed \modelname.

\section{Ethics Statement}
Our research focuses on automated long-text generation using LLMs, specifically for producing Wikipedia-style articles. While this technology holds great potential for enhancing knowledge accessibility and streamlining content creation, it also presents several ethical challenges that we acknowledge and actively address.

One key concern is that the generated content may delve into sensitive topics, as the model synthesizes information from diverse sources. Despite our extensive efforts to exclude controversial subjects during dataset construction, there remains a risk of producing biased, misleading, or inappropriate content. Such outputs could inadvertently activate LLM content safety mechanisms. We take this issue seriously and underscore the importance of advancing content filtering techniques and bias mitigation strategies to ensure the responsible use of this technology.

Another significant challenge lies in ensuring factual accuracy. Despite implementing measures to minimize errors, the model may still generate incorrect or hallucinated information. Worse still, if misused, this technology could actively contribute to the spread of false information, amplifying its potential societal impact. Addressing this issue calls for future research aimed at enhancing factual consistency, developing safeguards against misuse, and integrating more robust verification systems to validate the generated content.

\bibliography{main}

\begin{thebibliography}{43}
\providecommand{\natexlab}[1]{#1}

\bibitem[{Akbik et~al.(2019)Akbik, Bergmann, Blythe, Rasul, Schweter, and Vollgraf}]{akbik2019flair}
Alan Akbik, Tanja Bergmann, Duncan Blythe, Kashif Rasul, Stefan Schweter, and Roland Vollgraf. 2019.
\newblock {FLAIR}: An easy-to-use framework for state-of-the-art {NLP}.
\newblock In \emph{{NAACL} 2019, 2019 Annual Conference of the North American Chapter of the Association for Computational Linguistics (Demonstrations)}, pages 54--59.

\bibitem[{Asai et~al.(2023)Asai, Wu, Wang, Sil, and Hajishirzi}]{asai2023selfrag}
Akari Asai, Zeqiu Wu, Yizhong Wang, Avirup Sil, and Hannaneh Hajishirzi. 2023.
\newblock \href {https://arxiv.org/abs/2310.11511} {{Self-RAG}: Learning to retrieve, generate, and critique through self-reflection}.
\newblock \emph{arXiv preprint arXiv:2310.11511}.

\bibitem[{Bai et~al.(2024)Bai, Zhang, Lv, Zheng, Zhu, Hou, Dong, Tang, and Li}]{bai_longwriter_2024}
Yushi Bai, Jiajie Zhang, Xin Lv, Linzhi Zheng, Siqi Zhu, Lei Hou, Yuxiao Dong, Jie Tang, and Juanzi Li. 2024.
\newblock \href {http://arxiv.org/abs/2408.07055} {{LongWriter}: {Unleashing} 10,000+ {Word} {Generation} from {Long} {Context} {LLMs}}.
\newblock \emph{arXiv preprint}.
\newblock ArXiv:2408.07055 [cs].

\bibitem[{Balepur et~al.(2023)Balepur, Huang, and Chang}]{balepur-etal-2023-expository}
Nishant Balepur, Jie Huang, and Kevin Chang. 2023.
\newblock \href {https://doi.org/10.18653/v1/2023.emnlp-main.729} {Expository text generation: Imitate, retrieve, paraphrase}.
\newblock In \emph{Proceedings of the 2023 Conference on Empirical Methods in Natural Language Processing}, pages 11896--11919, Singapore. Association for Computational Linguistics.

\bibitem[{Chen et~al.(2024{\natexlab{a}})Chen, Cheng, Luu, and Bing}]{chen-etal-2024-exploring-potential}
Guizhen Chen, Liying Cheng, Anh~Tuan Luu, and Lidong Bing. 2024{\natexlab{a}}.
\newblock \href {https://doi.org/10.18653/v1/2024.acl-long.126} {Exploring the potential of large language models in computational argumentation}.
\newblock In \emph{Proceedings of the 62nd Annual Meeting of the Association for Computational Linguistics (Volume 1: Long Papers)}, pages 2309--2330, Bangkok, Thailand. Association for Computational Linguistics.

\bibitem[{Chen et~al.(2024{\natexlab{b}})Chen, Li, Zhang, and Zhang}]{chen-etal-2024-semantic}
Huiyao Chen, Xinxin Li, Meishan Zhang, and Min Zhang. 2024{\natexlab{b}}.
\newblock \href {https://doi.org/10.18653/v1/2024.findings-acl.527} {Semantic role labeling from {C}hinese speech via end-to-end learning}.
\newblock In \emph{Findings of the Association for Computational Linguistics ACL 2024}, pages 8898--8911, Bangkok, Thailand and virtual meeting. Association for Computational Linguistics.

\bibitem[{Doyle and Center(1994)}]{doyle1994information}
C.S. Doyle and Educational Resources~Information Center. 1994.
\newblock \href {https://books.google.co.jp/books?id=Z1IJ6A97WnsC} {\emph{Information Literacy in an Information Society: A Concept for the Information Age}}.
\newblock ERIC reports. DIANE Publishing.

\bibitem[{Edge et~al.(2024)Edge, Trinh, Cheng, Bradley, Chao, Mody, Truitt, and Larson}]{edge2024localglobalgraphrag}
Darren Edge, Ha~Trinh, Newman Cheng, Joshua Bradley, Alex Chao, Apurva Mody, Steven Truitt, and Jonathan Larson. 2024.
\newblock \href {https://arxiv.org/abs/2404.16130} {From local to global: A graph rag approach to query-focused summarization}.
\newblock \emph{Preprint}, arXiv:2404.16130.

\bibitem[{Fan and Gardent(2022)}]{fan-gardent-2022-generating}
Angela Fan and Claire Gardent. 2022.
\newblock \href {https://doi.org/10.18653/v1/2022.acl-long.586} {Generating biographies on {W}ikipedia: The impact of gender bias on the retrieval-based generation of women biographies}.
\newblock In \emph{Proceedings of the 60th Annual Meeting of the Association for Computational Linguistics (Volume 1: Long Papers)}, pages 8561--8576, Dublin, Ireland. Association for Computational Linguistics.

\bibitem[{Gao et~al.(2023)Gao, Ma, Lin, and Callan}]{gao-etal-2023-precise}
Luyu Gao, Xueguang Ma, Jimmy Lin, and Jamie Callan. 2023.
\newblock \href {https://doi.org/10.18653/v1/2023.acl-long.99} {Precise zero-shot dense retrieval without relevance labels}.
\newblock In \emph{Proceedings of the 61st Annual Meeting of the Association for Computational Linguistics (Volume 1: Long Papers)}, pages 1762--1777, Toronto, Canada. Association for Computational Linguistics.

\bibitem[{Guan et~al.(2021)Guan, Mao, Fan, Liu, Ding, and Huang}]{guan-etal-2021-long}
Jian Guan, Xiaoxi Mao, Changjie Fan, Zitao Liu, Wenbiao Ding, and Minlie Huang. 2021.
\newblock \href {https://doi.org/10.18653/v1/2021.acl-long.499} {Long text generation by modeling sentence-level and discourse-level coherence}.
\newblock In \emph{Proceedings of the 59th Annual Meeting of the Association for Computational Linguistics and the 11th International Joint Conference on Natural Language Processing (Volume 1: Long Papers)}, pages 6379--6393, Online. Association for Computational Linguistics.

\bibitem[{Hu et~al.(2024)Hu, Chen, Yang, Li, Zhang, Chen, and Chng}]{hu-etal-2024-gentranslate}
Yuchen Hu, Chen Chen, Chao-Han Yang, Ruizhe Li, Dong Zhang, Zhehuai Chen, and EngSiong Chng. 2024.
\newblock \href {https://doi.org/10.18653/v1/2024.acl-long.5} {{G}en{T}ranslate: Large language models are generative multilingual speech and machine translators}.
\newblock In \emph{Proceedings of the 62nd Annual Meeting of the Association for Computational Linguistics (Volume 1: Long Papers)}, pages 74--90, Bangkok, Thailand. Association for Computational Linguistics.

\bibitem[{Huang and Chen(2024)}]{huang2024factalignlongformfactualityalignment}
Chao-Wei Huang and Yun-Nung Chen. 2024.
\newblock \href {https://arxiv.org/abs/2410.01691} {Factalign: Long-form factuality alignment of large language models}.
\newblock \emph{Preprint}, arXiv:2410.01691.

\bibitem[{Jiang et~al.(2024{\natexlab{a}})Jiang, Shao, Ma, Semnani, and Lam}]{jiang2024unknownunknownsengagedhuman}
Yucheng Jiang, Yijia Shao, Dekun Ma, Sina~J. Semnani, and Monica~S. Lam. 2024{\natexlab{a}}.
\newblock \href {https://arxiv.org/abs/2408.15232} {Into the unknown unknowns: Engaged human learning through participation in language model agent conversations}.
\newblock \emph{Preprint}, arXiv:2408.15232.

\bibitem[{Jiang et~al.(2024{\natexlab{b}})Jiang, Ma, and Chen}]{jiang2024longragenhancingretrievalaugmentedgeneration}
Ziyan Jiang, Xueguang Ma, and Wenhu Chen. 2024{\natexlab{b}}.
\newblock \href {https://arxiv.org/abs/2406.15319} {Longrag: Enhancing retrieval-augmented generation with long-context llms}.
\newblock \emph{Preprint}, arXiv:2406.15319.

\bibitem[{Joshi et~al.(2017)Joshi, Choi, Weld, and Zettlemoyer}]{joshi-etal-2017-triviaqa}
Mandar Joshi, Eunsol Choi, Daniel Weld, and Luke Zettlemoyer. 2017.
\newblock \href {https://doi.org/10.18653/v1/P17-1147} {{T}rivia{QA}: A large scale distantly supervised challenge dataset for reading comprehension}.
\newblock In \emph{Proceedings of the 55th Annual Meeting of the Association for Computational Linguistics (Volume 1: Long Papers)}, pages 1601--1611, Vancouver, Canada. Association for Computational Linguistics.

\bibitem[{Kang and Xiong(2024)}]{kang_researcharena_2024}
Hao Kang and Chenyan Xiong. 2024.
\newblock \href {https://doi.org/10.48550/arXiv.2406.10291} {{ResearchArena}: {Benchmarking} {LLMs}' {Ability} to {Collect} and {Organize} {Information} as {Research} {Agents}}.
\newblock \emph{arXiv preprint}.
\newblock ArXiv:2406.10291 [cs].

\bibitem[{Kim et~al.(2024)Kim, Suk, Longpre, Lin, Shin, Welleck, Neubig, Lee, Lee, and Seo}]{kim2024prometheus}
Seungone Kim, Juyoung Suk, Shayne Longpre, Bill~Yuchen Lin, Jamin Shin, Sean Welleck, Graham Neubig, Moontae Lee, Kyungjae Lee, and Minjoon Seo. 2024.
\newblock \href {https://arxiv.org/abs/2405.01535} {Prometheus 2: An open source language model specialized in evaluating other language models}.
\newblock \emph{Preprint}, arXiv:2405.01535.

\bibitem[{Lei et~al.(2024)Lei, Guo, He, Zhang, Zhang, Peng, Liu, and Chen}]{lei_ex3_2024}
Huang Lei, Jiaming Guo, Guanhua He, Xishan Zhang, Rui Zhang, Shaohui Peng, Shaoli Liu, and Tianshi Chen. 2024.
\newblock \href {http://arxiv.org/abs/2408.08506} {Ex3: {Automatic} {Novel} {Writing} by {Extracting}, {Excelsior} and {Expanding}}.
\newblock \emph{arXiv preprint}.
\newblock ArXiv:2408.08506 [cs].

\bibitem[{Lewis et~al.(2020)Lewis, Perez, Piktus, Petroni, Karpukhin, Goyal, K{\"u}ttler, Lewis, Yih, Rockt{\"a}schel et~al.}]{lewis2020retrieval}
Patrick Lewis, Ethan Perez, Aleksandra Piktus, Fabio Petroni, Vladimir Karpukhin, Naman Goyal, Heinrich K{\"u}ttler, Mike Lewis, Wen-tau Yih, Tim Rockt{\"a}schel, et~al. 2020.
\newblock Retrieval-augmented generation for knowledge-intensive nlp tasks.
\newblock \emph{Advances in Neural Information Processing Systems}, 33:9459--9474.

\bibitem[{Li et~al.(2024)Li, Chen, Yan, Wang, Zhang, and Sundaram}]{li-etal-2024-advancing}
Yunzhe Li, Qian Chen, Weixiang Yan, Wen Wang, Qinglin Zhang, and Hari Sundaram. 2024.
\newblock \href {https://aclanthology.org/2024.eacl-long.145/} {Advancing precise outline-conditioned text generation with task duality and explicit outline control}.
\newblock In \emph{Proceedings of the 18th Conference of the European Chapter of the Association for Computational Linguistics (Volume 1: Long Papers)}, pages 2362--2377, St. Julian{'}s, Malta. Association for Computational Linguistics.

\bibitem[{Lin(2004)}]{lin-2004-rouge}
Chin-Yew Lin. 2004.
\newblock \href {https://aclanthology.org/W04-1013} {{ROUGE}: A package for automatic evaluation of summaries}.
\newblock In \emph{Text Summarization Branches Out}, pages 74--81, Barcelona, Spain. Association for Computational Linguistics.

\bibitem[{Liu et~al.(2024{\natexlab{a}})Liu, Zhang, Guo, Wang, Dong, Li, Lee, Zhang, and Liu}]{liu2024ctrlaadaptiveretrievalaugmentedgeneration}
Huanshuo Liu, Hao Zhang, Zhijiang Guo, Jing Wang, Kuicai Dong, Xiangyang Li, Yi~Quan Lee, Cong Zhang, and Yong Liu. 2024{\natexlab{a}}.
\newblock \href {https://arxiv.org/abs/2405.18727} {Ctrla: Adaptive retrieval-augmented generation via inherent control}.
\newblock \emph{Preprint}, arXiv:2405.18727.

\bibitem[{Liu et~al.(2024{\natexlab{b}})Liu, Shi, He, Ye, Fabbri, Liu, Radev, and Cohan}]{liu-etal-2024-learning}
Yixin Liu, Kejian Shi, Katherine He, Longtian Ye, Alexander Fabbri, Pengfei Liu, Dragomir Radev, and Arman Cohan. 2024{\natexlab{b}}.
\newblock \href {https://doi.org/10.18653/v1/2024.naacl-long.478} {On learning to summarize with large language models as references}.
\newblock In \emph{Proceedings of the 2024 Conference of the North American Chapter of the Association for Computational Linguistics: Human Language Technologies (Volume 1: Long Papers)}, pages 8647--8664, Mexico City, Mexico. Association for Computational Linguistics.

\bibitem[{Ma et~al.(2023)Ma, Gong, He, Zhao, and Duan}]{ma-etal-2023-query}
Xinbei Ma, Yeyun Gong, Pengcheng He, Hai Zhao, and Nan Duan. 2023.
\newblock \href {https://doi.org/10.18653/v1/2023.emnlp-main.322} {Query rewriting in retrieval-augmented large language models}.
\newblock In \emph{Proceedings of the 2023 Conference on Empirical Methods in Natural Language Processing}, pages 5303--5315, Singapore. Association for Computational Linguistics.

\bibitem[{Mallen et~al.(2023)Mallen, Asai, Zhong, Das, Khashabi, and Hajishirzi}]{mallen-etal-2023-trust}
Alex Mallen, Akari Asai, Victor Zhong, Rajarshi Das, Daniel Khashabi, and Hannaneh Hajishirzi. 2023.
\newblock \href {https://doi.org/10.18653/v1/2023.acl-long.546} {When not to trust language models: Investigating effectiveness of parametric and non-parametric memories}.
\newblock In \emph{Proceedings of the 61st Annual Meeting of the Association for Computational Linguistics (Volume 1: Long Papers)}, pages 9802--9822, Toronto, Canada. Association for Computational Linguistics.

\bibitem[{Min et~al.(2023)Min, Krishna, Lyu, Lewis, Yih, Koh, Iyyer, Zettlemoyer, and Hajishirzi}]{min-etal-2023-factscore}
Sewon Min, Kalpesh Krishna, Xinxi Lyu, Mike Lewis, Wen-tau Yih, Pang Koh, Mohit Iyyer, Luke Zettlemoyer, and Hannaneh Hajishirzi. 2023.
\newblock \href {https://doi.org/10.18653/v1/2023.emnlp-main.741} {{FA}ct{S}core: Fine-grained atomic evaluation of factual precision in long form text generation}.
\newblock In \emph{Proceedings of the 2023 Conference on Empirical Methods in Natural Language Processing}, pages 12076--12100, Singapore. Association for Computational Linguistics.

\bibitem[{Pham et~al.(2024)Pham, Sun, and Iyyer}]{pham2024surimulticonstraintinstructionfollowing}
Chau~Minh Pham, Simeng Sun, and Mohit Iyyer. 2024.
\newblock \href {https://arxiv.org/abs/2406.19371} {Suri: Multi-constraint instruction following for long-form text generation}.
\newblock \emph{Preprint}, arXiv:2406.19371.

\bibitem[{Rawte et~al.(2023)Rawte, Chakraborty, Pathak, Sarkar, Tonmoy, Chadha, Sheth, and Das}]{rawte-etal-2023-troubling}
Vipula Rawte, Swagata Chakraborty, Agnibh Pathak, Anubhav Sarkar, S.M Towhidul~Islam Tonmoy, Aman Chadha, Amit Sheth, and Amitava Das. 2023.
\newblock \href {https://doi.org/10.18653/v1/2023.emnlp-main.155} {The troubling emergence of hallucination in large language models - an extensive definition, quantification, and prescriptive remediations}.
\newblock In \emph{Proceedings of the 2023 Conference on Empirical Methods in Natural Language Processing}, pages 2541--2573, Singapore. Association for Computational Linguistics.

\bibitem[{Reimers and Gurevych(2019)}]{reimers-2019-sentence-bert}
Nils Reimers and Iryna Gurevych. 2019.
\newblock \href {http://arxiv.org/abs/1908.10084} {Sentence-bert: Sentence embeddings using siamese bert-networks}.
\newblock In \emph{Proceedings of the 2019 Conference on Empirical Methods in Natural Language Processing}. Association for Computational Linguistics.

\bibitem[{Rohman(1965)}]{prewriting-stage}
D.~Gordon Rohman. 1965.
\newblock \href {http://www.jstor.org/stable/354885} {Pre-writing the stage of discovery in the writing process}.
\newblock \emph{College Composition and Communication}, (2):106--112.

\bibitem[{Shao et~al.(2024)Shao, Jiang, Kanell, Xu, Khattab, and Lam}]{shao-etal-2024-assisting}
Yijia Shao, Yucheng Jiang, Theodore Kanell, Peter Xu, Omar Khattab, and Monica Lam. 2024.
\newblock \href {https://doi.org/10.18653/v1/2024.naacl-long.347} {Assisting in writing {W}ikipedia-like articles from scratch with large language models}.
\newblock In \emph{Proceedings of the 2024 Conference of the North American Chapter of the Association for Computational Linguistics: Human Language Technologies (Volume 1: Long Papers)}, pages 6252--6278, Mexico City, Mexico. Association for Computational Linguistics.

\bibitem[{Shen et~al.(2023)Shen, August, Siangliulue, Lo, Bragg, Hammerbacher, Downey, Chang, and Sontag}]{shen2023summarizationdesigningaisupport}
Zejiang Shen, Tal August, Pao Siangliulue, Kyle Lo, Jonathan Bragg, Jeff Hammerbacher, Doug Downey, Joseph~Chee Chang, and David Sontag. 2023.
\newblock \href {https://arxiv.org/abs/2304.02623} {Beyond summarization: Designing ai support for real-world expository writing tasks}.
\newblock \emph{Preprint}, arXiv:2304.02623.

\bibitem[{Tan et~al.(2021)Tan, Yang, Al-Shedivat, Xing, and Hu}]{tan-etal-2021-progressive}
Bowen Tan, Zichao Yang, Maruan Al-Shedivat, Eric Xing, and Zhiting Hu. 2021.
\newblock \href {https://doi.org/10.18653/v1/2021.naacl-main.341} {Progressive generation of long text with pretrained language models}.
\newblock In \emph{Proceedings of the 2021 Conference of the North American Chapter of the Association for Computational Linguistics: Human Language Technologies}, pages 4313--4324, Online. Association for Computational Linguistics.

\bibitem[{Wang et~al.(2022)Wang, Yang, Huang, Jiao, Yang, Jiang, Majumder, and Wei}]{wang2022text}
Liang Wang, Nan Yang, Xiaolong Huang, Binxing Jiao, Linjun Yang, Daxin Jiang, Rangan Majumder, and Furu Wei. 2022.
\newblock \href {https://doi.org/10.48550/arXiv.2212.03533} {Text embeddings by weakly-supervised contrastive pre-training}.
\newblock \emph{arXiv preprint arXiv:2212.03533}.

\bibitem[{Wang et~al.(2024)Wang, Guo, Yao, Zhang, Zhang, Wu, Zhang, Dai, Zhang, Wen, Ye, Zhang, and Zhang}]{wang2024autosurveylargelanguagemodels}
Yidong Wang, Qi~Guo, Wenjin Yao, Hongbo Zhang, Xin Zhang, Zhen Wu, Meishan Zhang, Xinyu Dai, Min Zhang, Qingsong Wen, Wei Ye, Shikun Zhang, and Yue Zhang. 2024.
\newblock \href {https://arxiv.org/abs/2406.10252} {Autosurvey: Large language models can automatically write surveys}.
\newblock \emph{Preprint}, arXiv:2406.10252.

\bibitem[{Xu et~al.(2024)Xu, Ping, Wu, Xu, Liu, Shoeybi, and Catanzaro}]{xu2024chatqa2bridginggap}
Peng Xu, Wei Ping, Xianchao Wu, Chejian Xu, Zihan Liu, Mohammad Shoeybi, and Bryan Catanzaro. 2024.
\newblock \href {https://arxiv.org/abs/2407.14482} {Chatqa 2: Bridging the gap to proprietary llms in long context and rag capabilities}.
\newblock \emph{Preprint}, arXiv:2407.14482.

\bibitem[{Yang et~al.(2022)Yang, Liu, Xiong, Zhang, Chen, and Xu}]{yang-etal-2022-long}
Erguang Yang, Mingtong Liu, Deyi Xiong, Yujie Zhang, Yufeng Chen, and Jinan Xu. 2022.
\newblock \href {https://doi.org/10.18653/v1/2022.emnlp-main.554} {Long text generation with topic-aware discrete latent variable model}.
\newblock In \emph{Proceedings of the 2022 Conference on Empirical Methods in Natural Language Processing}, pages 8100--8107, Abu Dhabi, United Arab Emirates. Association for Computational Linguistics.

\bibitem[{Yang et~al.(2023)Yang, Klein, Peng, and Tian}]{yang-etal-2023-doc}
Kevin Yang, Dan Klein, Nanyun Peng, and Yuandong Tian. 2023.
\newblock \href {https://doi.org/10.18653/v1/2023.acl-long.190} {{DOC}: Improving long story coherence with detailed outline control}.
\newblock In \emph{Proceedings of the 61st Annual Meeting of the Association for Computational Linguistics (Volume 1: Long Papers)}, pages 3378--3465, Toronto, Canada. Association for Computational Linguistics.

\bibitem[{Yang et~al.(2018)Yang, Qi, Zhang, Bengio, Cohen, Salakhutdinov, and Manning}]{yang2018hotpotqa}
Zhilin Yang, Peng Qi, Saizheng Zhang, Yoshua Bengio, William~W. Cohen, Ruslan Salakhutdinov, and Christopher~D. Manning. 2018.
\newblock {HotpotQA}: A dataset for diverse, explainable multi-hop question answering.
\newblock In \emph{Conference on Empirical Methods in Natural Language Processing ({EMNLP})}.

\bibitem[{You et~al.(2023)You, Wu, Liang, Mao, Wu, Cao, Cai, Guo, Xia, Wei, and Duan}]{you2023eipetextevaluationguidediterativeplan}
Wang You, Wenshan Wu, Yaobo Liang, Shaoguang Mao, Chenfei Wu, Maosong Cao, Yuzhe Cai, Yiduo Guo, Yan Xia, Furu Wei, and Nan Duan. 2023.
\newblock \href {https://arxiv.org/abs/2310.08185} {Eipe-text: Evaluation-guided iterative plan extraction for long-form narrative text generation}.
\newblock \emph{Preprint}, arXiv:2310.08185.

\bibitem[{Yu et~al.(2024)Yu, Zang, Wang, Zhuang, and Gu}]{yu-etal-2024-charpoet}
Chengyue Yu, Lei Zang, Jiaotuan Wang, Chenyi Zhuang, and Jinjie Gu. 2024.
\newblock \href {https://doi.org/10.18653/v1/2024.acl-demos.30} {{C}har{P}oet: A {C}hinese classical poetry generation system based on token-free {LLM}}.
\newblock In \emph{Proceedings of the 62nd Annual Meeting of the Association for Computational Linguistics (Volume 3: System Demonstrations)}, pages 315--325, Bangkok, Thailand. Association for Computational Linguistics.

\bibitem[{Zhou et~al.(2023)Zhou, Jiang, Cui, Wang, Xiao, Hou, Cotterell, and Sachan}]{zhou_recurrentgpt_2023}
Wangchunshu Zhou, Yuchen~Eleanor Jiang, Peng Cui, Tiannan Wang, Zhenxin Xiao, Yifan Hou, Ryan Cotterell, and Mrinmaya Sachan. 2023.
\newblock \href {https://doi.org/10.48550/arXiv.2305.13304} {{RecurrentGPT}: {Interactive} {Generation} of ({Arbitrarily}) {Long} {Text}}.
\newblock \emph{arXiv preprint}.
\newblock ArXiv:2305.13304 [cs].

\end{thebibliography}

\appendix

\newpage
\section{Dataset Details}
\label{app:dataset_stats}

\begin{figure}[h]
  \centering
  \includegraphics[width=1\linewidth]{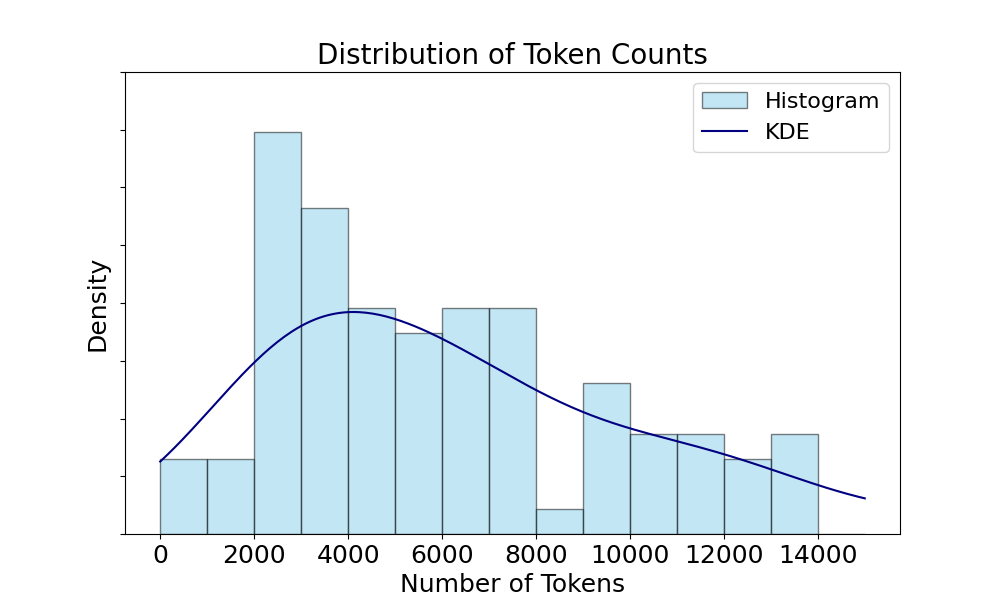}
  \caption{The distribution of the length of FreshWiki-2024}
  \label{fig:len_distribution}
\end{figure}

We adopted the data collection methodology outlined in FreshWiki to construct our dataset. Specifically, we systematically crawled the top 100 most frequently edited topics monthly from January to December 2024, retaining only articles rated as Class B or higher by the ORES quality assessment system\footnote{https://www.mediawiki.org/wiki/ORES}. To streamline processing and focus our analysis, we removed all tabular data and multimodal content, preserving only plain text following the same approach as STORM.

From this curated collection, we randomly selected 100 topics for model evaluation. As shown in Figure~\ref{fig:len_distribution}, article lengths followed a bimodal distribution: approximately 70\% of articles ranged between 2,000 and 8,000 tokens, while we intentionally retained articles exceeding 15,000 tokens to evaluate the model's performance or processing intricate topic matter.

\begin{figure}[h]
  \centering
  \includegraphics[width=1\linewidth]{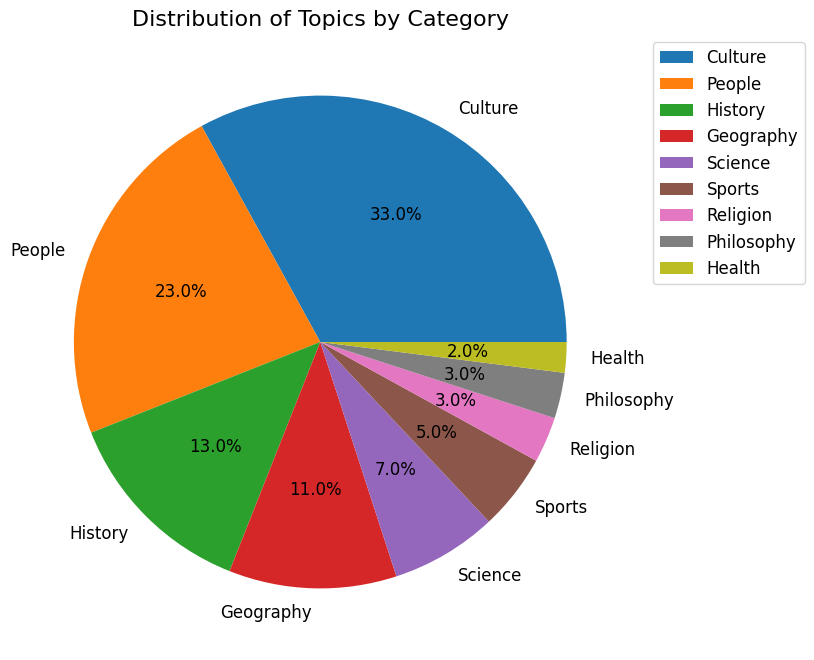}
  \caption{The distribution of classification of Freshwiki-2024 }
  \label{fig:topic_distribution}
\end{figure}

To ensure thematic diversity, we categorized the 100 selected topics into four distinct domains using GPT-4o. Figure~\ref{fig:topic_distribution} illustrates the resulting category distribution, demonstrating broad coverage and relatively balanced representation across various subject areas.

\section{Evaluation Details}

\subsection{Limitations for NER-based outline evaluation methods}
\label{limitations_NER}
STORM's NER-based evaluation metrics, namely Entity Recall and Soft Entity Recall, exhibit two fundamental limitations. First, conventional NER models are not specifically designed for outline evaluation. In particular, key section titles (e.g., "Background") in outlines often do not correspond to well-defined named entities. As a result, extracting entities from golden outlines can be unreliable, and in extreme cases, no identifiable entities may be present at all. Second, the soft entity recall metric suffers from abnormal scaling behavior due to the lack of a normalization mechanism. This absence of normalization, coupled with the inherent limitations of entity recognition, can lead to inflated scores that misrepresent the actual quality of an outline.

\subsection{Article Quality Evaluation}
\label{article_evaluation}
We utilize the Qwen-Turbo model as the backbone for FactScore, which decomposes atomic facts and assesses their accuracy. FactScore can be regarded as a measure of the factual accuracy of generated text. However, comparing FactScore across different texts can be challenging when the generated content varies significantly in length. To address this issue, we utilize FactAlign, which assumes a predefined number of claims, denoted as $K$, for each article. The recall is then computed as the ratio of generated claims to $K$, allowing for a more balanced evaluation. Based on this, we calculate the $F1@K$ score, where we set $K=300$ in our experiments. Regarding LLM evaluation, we observed that due to the context length limitations of \texttt{Prometheus-7b-v1.0}, STORM processes all generated articles to a limited length of 2,000 words, which significantly impacts the fairness of the evaluation. Thanks to the extended context length of \texttt{Prometheus-7b-v2.0}, we were able to process articles up to 10,000 words in our experiments, leading to a more comprehensive assessment. The detailed prompts for LLM Evaluation are shown in Table~\ref{table:llm_rubric}. Table~\ref{tab:pearson-correlation} shows the Pearson correlation coefficient data between human evaluation and LLM evaluation. Additionally, we utilized \texttt{text-embedding-3-small} to generate embeddings for the info diversity metric, following the implementation of CoSTORM.
\begin{table}[t]
\centering

\begin{tabular}{l|c} 
\toprule
          & \textbf{Pearson Correlation} \\ 
\midrule
          Interst Level & 0.58 \\
          Organization & 0.51 \\
          Relevance & 0.31 \\
          Coverage & 0.47 \\
\bottomrule
\end{tabular}
\caption{Pearson correlation between average human evaluation scores and LLM-based evaluation scores on genreated article quality (n=20)}
\label{tab:pearson-correlation}
\end{table}

\begin{table*}
\centering
\scriptsize
\begin{tblr}{
    hline{1,2,7,8,13,14,20,21,26} = {-}{0.08em},
    colspec = {ll}
}

Criteria Description & \textbf{Interest Level}: How engaging and thought-provoking is the article?\\
Score 1 Description & Not engaging at all; no attempt to capture the reader’s attention.\\
Score 2 Description & Fairly engaging with a basic narrative but lacking depth.\\
Score 3 Description & Moderately engaging with several interesting points.\\
Score 4 Description & Quite engaging with a well-structured narrative and noteworthy points that frequently capture and retain attention.\\
Score 5 Description & Exceptionally engaging throughout, with a compelling narrative that consistently stimulates interest.\\

Criteria Description & \textbf{Coherence and Organization}: Is the article well-organized and logically structured?\\
Score 1 Description & Disorganized; lacks logical structure and coherence.\\
Score 2 Description & Fairly organized; a basic structure is present but not consistently followed.\\
Score 3 Description & Organized; a clear structure is mostly followed with some lapses in coherence.\\
Score 4 Description & Good organization; a clear structure with minor lapses in coherence.\\
Score 5 Description & Excellently organized; the article is logically structured with seamless transitions and a clear argument.\\

Criteria Description & \textbf{Relevance and Focus}: Does the article stay on topic and maintain a clear focus?\\
Score 1 Description & Off-topic; the content does not align with the headline or core subject.\\
Score 2 Description & Somewhat on topic but with several digressions; the core subject is evident but not consistently adhered to.\\
Score 3 Description & Generally on topic, despite a few unrelated details.\\
Score 4 Description & Mostly on topic and focused; the narrative has a consistent relevance to the core subject with infrequent digressions.\\
Score 5 Description & Exceptionally focused and entirely on topic; the article is tightly centered on the subject, with every piece of inform- \\
& ation contributing to a comprehensive understanding of the topic.\\

Criteria Description & \textbf{Broad Coverage}: Does the article provide an in-depth exploration of the topic and have good coverage?\\
Score 1 Description & Severely lacking; offers little to no coverage of the topic's primary aspects, resulting in a very narrow perspective.\\
Score 2 Description & Partial coverage; includes some of the topic's main aspects but misses others, resulting in an incomplete portrayal.\\
Score 3 Description & Acceptable breadth; covers most main aspects, though it may stray into minor unnecessary details or overlook some relevant points.\\
Score 4 Description & Good coverage; achieves broad coverage of the topic, hitting on all major points with minimal extraneous information.\\
Score 5 Description & Exemplary in breadth; delivers outstanding coverage, thoroughly detailing all crucial aspects of the topic without including irrelevant information.
                 
\end{tblr}
\caption{Scoring rubrics for LLM Evaluation}
\label{table:llm_rubric}
\end{table*}

\section{Full Prompt}
\label{sec:prompt}

As discussed in \S\ref{section:methods}, we divided our \modelname framework into three modules and listed the prompts used for each. The prompts we use for \S\S\ref{sec:outline_gen} are listed in Figure \ref{prompt:topic_summarization}, \ref{prompt:rag_outline_generation} while \S\S\ref{sec:info_seek} in Figure \ref{prompt:attributes extraction}, \ref{prompt:attri2query}, \ref{prompt:queries_merging}, \ref{prompt:operation_generation}, \ref{prompt:outline_refinement} and \S\S\ref{sec:article_gen} in Figure~\ref{prompt:plan_generation}, \ref{prompt:section_writing}.

\section{Efficiency and Usage Analysis}
\label{Efficiency_Analysis}

\begin{figure}[t]
  \centering
  \includegraphics[width=1.0\linewidth]{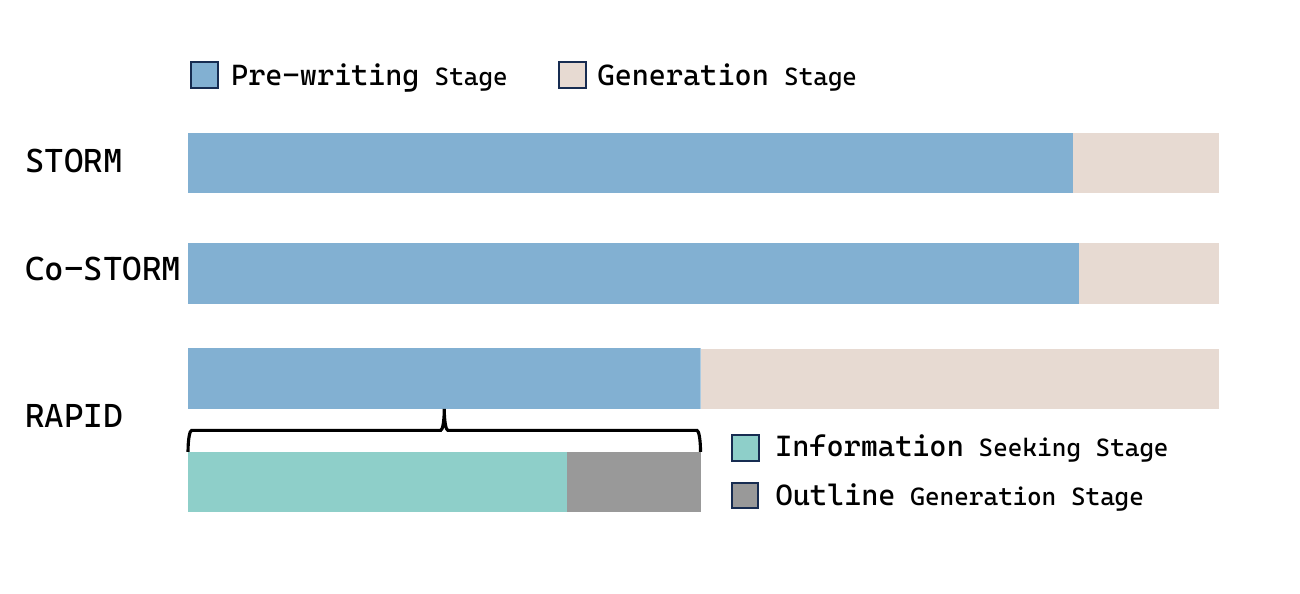}
    
  \caption{The distribution of time consumed in each stage of the pipelines.}

  \label{fig:time}
\end{figure}

\modelname~integrates an attribute-constrained information-seeking module, which is designed to enhance efficiency while maintaining comprehensiveness. To evaluate its efficiency and usability compared to other baselines, we collected the average time and API usage for generating 100 topics. For methods involving parallel section generation, we set the maximum parallelism to 3. As shown in the table, although the first three simple baselines consume less time and resources, they struggle to generate reliable articles. In contrast, \modelname~not only produces high-quality articles but also significantly reduces API usage and time consumption compared to STORM and Co-STORM.
\begin{table}[t]
  \centering
  \small
  \begin{tblr}{
    hline{1,2,4,6,7} = {-}{0.08em},
    colspec = {lccc}
  }

          & \textbf{Calls(/it)} & \textbf{Tokens(k/it)}   & \textbf{Time(s/it)}    \\
  RAG     &2 &6.37  &32.12 \\
  ORAG    &14.25 &24.13  &109.83 \\
  STORM   & 88.06 & 60.71 & 163.22  \\
  Co-STORM & 70.30  & 50.52 & 154.14  \\
  \textbf{\modelname}   &\textbf{31.04} & \textbf{43.62} & \textbf{127.19}

  \end{tblr}
  \caption{Results of pipeline efficiency evaluation. We evaluated the average API calls, token consumption, and the total duration time of generating an article. }
  \label{efficiency_result}
\end{table}
Additionally, we analyzed the proportion of time consumed at each stage by STORM, Co-STORM, and \modelname, as illustrated in Figure~\ref{fig:time}. The results indicate that both STORM and Co-STORM predominantly allocate a substantial amount of time to the pre-writing stage, leading to significant time consumption. This is largely due to their reliance on an agent-based dialogue mechanism during the pre-writing phase, which is inherently difficult to optimize further. In contrast, \modelname~drastically reduces the time spent in the pre-writing stage through its parallelized information collection approach, enabling more efficient resource allocation across all stages.

\section{Case Study}
\label{case_study}
We utilized \modelname to generate an article on the topic of AlphaFold. The complete outline and writing plan are illustrated in Figures~\ref{case_outline} and \ref{case_plan}, respectively. Due to space limitations, only a small portion of the final generated article is presented in Figure \ref{case_article}. This case demonstrates that \modelname is capable of effectively generating a coherent writing plan and producing a comprehensive and consistent long-form article.

\begin{figure*}

\begin{tcolorbox}[
colback=gray!10!white,
colframe=black!75!black,
colbacktitle=black!60!white, 
title=Topic Summarization,  
width=\textwidth, 
]

You are an expert in utilizing search engines effectively. You are currently compiling information for a wiki page based on a given topic. Now that you have obtained content returned by search engines regarding this topic, please analyze whether there are any ambiguities or multiple concepts associated with it. If the topic is clear, generate a brief introduction based on the search engine content to clarify the concept for subsequent writing, ensuring that the introduction remains within three sentences.\\

Topic you are discussing about: <topic>\\

Gathered information from search engines: <search results>\\

Now give your response. Make sure that only the introduction is outputted. Do not repeat the input prompt.

\end{tcolorbox}
\caption{prompt used for Topic Summarization}
\label{prompt:topic_summarization}
\end{figure*}
\begin{figure*}

\begin{tcolorbox}[
colback=gray!10!white,
colframe=black!75!black,
colbacktitle=black!60!white, 
title=RAG Outline Generation,  
width=\textwidth, 
]
Write an outline for a Wikipedia page. \\
Here is the format of your writing: \\
1. Use "\#" Title" to indicate section title, "\#\#" Title" to indicate subsection title, "\#\#\#" Title" to indicate subsubsection title, and so on. \\
2. Do not include other information. \\
3. Do not include topic name itself in the outline. \\

The topic you want to write: <topic> \\

Brief intro of the topic: <brief info>\\ 

Outlines of similar topics: <similar topic outline>\\

Write the outline of the topic: 

\end{tcolorbox}
\caption{Prompt used for RAG Outline Generation}
\label{prompt:rag_outline_generation}
\end{figure*}
\begin{figure*}

\begin{tcolorbox}[
colback=gray!10!white,
colframe=black!75!black,
colbacktitle=black!60!white, 
title=Attributes Extraction,  
width=\textwidth, 
]
Generate attributes for a specified topic with its outline. The generated attributes should summarize all information needed to write the wiki page for this topic. Please avoid creating complex attributes; ensure that each attribute represents a distinct and indivisible aspect. \\
Here is the format of your response: \\
1. Generate attributes, each on a new line, ensuring no additional tags or formatting are included. \\
2. Do not include other information.\\ 
3. Do not include topic name itself in the attribute list.\\

Topic: <topic> \\

Outline: <outline> \\

Attributes:

\end{tcolorbox}
\caption{Prompt used for Attributes Extraction}
\label{prompt:attributes extraction}
\end{figure*}
\begin{figure*}

\begin{tcolorbox}[
colback=gray!10!white,
colframe=black!75!black,
colbacktitle=black!60!white, 
title=Attributes to Queries,  
width=\textwidth, 
]
You want to search the info of attributes of the topic using Google search. What do you type in the search box? Write the queries you will use in the following format: \\
- query 1 \\ 
- query 2 \\
- ...   \\ 
- query n\\

Topic you are discussing about: <topic> \\

The attributes of the topic: <attributes> \\

Now give your response. Make sure that only queries are output. Do not repeat the input prompt

\end{tcolorbox}
\caption{Prompt used for Attributes to Queries}
\label{prompt:attri2query}
\end{figure*}
\begin{figure*}

\begin{tcolorbox}[
colback=gray!10!white,
colframe=black!75!black,
colbacktitle=black!60!white, 
title=Queries Merging,  
width=\textwidth, 
]
I want you to act as a researcher gathering information to compose a wiki article on a specific topic. You are now presented with a topic and a list of queries designed to gather information for the topic. Your task is to modify or enhance the query list based on the relevant topics and their queries. Ensure that the final queries comprehensively encompass information beneficial for writing about the topic and are suitable for use in Google searches. Do not repeat the input prompt. \\Here is the format of your response:\\ 
- query 1 \\ 
- query 2  \\
- ...  \\
- query n \\ 

The topic you are discussing about: <topic> \\
 
The queries of the topic: <queries> \\

The similar topics with their queries: <similar topics with queries> \\
 
The final response of the queries: 

\end{tcolorbox}
\caption{Prompt used for Queries Merging}
\label{prompt:queries_merging}
\end{figure*}
\begin{figure*}

\begin{tcolorbox}[
colback=gray!10!white,
colframe=black!75!black,
colbacktitle=black!60!white, 
title=Operation Generation,  
width=\textwidth, 
]
You are improving an outline for a wiki page. Now I will give you a draft outline and some titles of the searched results. You can do three operations:\\ 
add section \\
delete section\\ 
do nothing \\

Please list the operations you need to do:\newline 
-[add section] : section\_title \newline
-[delete section] : section\_title\newline
If nothing is needed to do, please just generate:\newline 
-[do nothing] \\

Directly write the operations and do not include any other information.\\

The topic you want to write: <topic>\\

The draft outline: <outline> \\

Titles of the searched results: <titles>\\

Please generate the operations:

\end{tcolorbox}
\caption{Prompt used for Operation Generation}
\label{prompt:operation_generation}
\end{figure*}
\begin{figure*}

\begin{tcolorbox}[
colback=gray!10!white,
colframe=black!75!black,
colbacktitle=black!60!white, 
title=Outline Refinement,  
width=\textwidth, 
]
You are improving an outline for a wiki page. Now I will give you a draft outline and some operations like: \newline
-[add section] \newline
-[delete section] \newline
-[do nothing] 

Please proceed with the operations for the outline and then refine the overall outline. Directly write the refined outline and do not include any other information. \\

The topic you want to write: <topic> \\

The draft outline: <outline>\\

The operations: <operations>\\

Please generate the refined outline:

\end{tcolorbox}
\caption{Prompt used for Outline Refinement}
\label{prompt:outline_refinement}
\end{figure*}
\begin{figure*}

\begin{tcolorbox}[
colback=gray!10!white,
colframe=black!75!black,
colbacktitle=black!60!white, 
title=Plan Generation,  
width=\textwidth, 
]
You are an experienced wiki writer. I will provide you with a topic with its outline to write. I want you to generate a writing plan for this outline to improve the coherence of the article. The plan defines which sections is needed to be generated before the current section. Try to choose the sections that can help improve the coherence and fluency of the current section. For example, sections like 'Background' don't need extra information while sections like 'Introduction' or 'Conclusion' need all other sections. Please just generate the plan for the first level sections and make sure that the plan is in a valid topological order. If no extra information is needed, generate "None". All the needed sections are connected by '<-' and make sure that they are all from the first level sections of outline. Just output the plan and do not explain. \\

Here is an example: <example> \\

Topic: <topic> \\

Outline: <outline>\\ 

Generate the plan of the given topic and outline(do not repeat the outline):

\end{tcolorbox}
\caption{Prompt used for Plan Generation}
\label{prompt:plan_generation}
\end{figure*}
\begin{figure*}

\begin{tcolorbox}[
colback=gray!10!white,
colframe=black!75!black,
colbacktitle=black!60!white, 
title=Section Writing,  
width=\textwidth, 
]

Write a Wikipedia section based on the collected information. Here is the format of your writing: \\
1. Use "\#" Title" to indicate section title, "\#\#" Title" to indicate subsection title, "\#\#\#" Title" to indicate subsubsection title, and so on.\\ 
2. Use [1], [2], ..., [n] in line (for example, "The capital of the United States is Washington, D.C.[1][3]."). \\
You DO NOT need to include a References or Sources section to list the sources at the end. 

The collected information: <collected info> \\
 
The topic of the page: <topic>\\

The other sections of the page: <other sections> \\

The section you need to write: <section title> \\
 
The outline of the section: <section outline> \\
 
Write the section with proper inline citations (Start your writing with \# section title. Don't include the page title or try to write other sections):

\end{tcolorbox}
\caption{prompt used for Section Writing}
\label{prompt:section_writing}
\end{figure*}

\begin{figure*}

\begin{tcolorbox}[
colback=gray!10!white,
colframe=black!75!black,
colbacktitle=black!60!white, 
title=AlphaFold,  
width=\textwidth, 
]
\# Background

\# Development

\#\# Origin of the project

\#\# Collaboration with scientific communities

\# Algorithm

\#\# Neural network architecture

\#\# Training datasets

\#\# AlphaFold 1

\#\# AlphaFold 2

\#\# Updates and enhancements in AlphaFold 3

\#\# AlphaFold 3 Pairformer and advancements

\# Protein Structure Predictions

\#\# Methodology

\#\# Accuracy and reliability

\#\# Comparison with experimental methods

\# Competitions and Benchmarks

\#\# CASP12

\#\# CASP13

\#\# CASP14

\#\# CASP15

\# Applications

\#\# Drug discovery

\#\# Enzyme engineering

\#\# Understanding diseases

\#\# Environmental research

\#\# Synthetic biology

\#\# Role in antibody characterization

\#\# Use of AlphaFold in predicting genetic mutation impacts

\# Database

\#\# AlphaFold Protein Structure Database

\#\# Accessibility and usage

\#\# Global impact on research

\# Source Code and Open Access

\#\# Release of AlphaFold source code

\#\# Community contributions and forks

\# Limitations

\#\# Known issues

\#\# Challenges in complex protein assemblies

\#\# Limitations with membrane proteins and disordered regions

\#\# Limitations in predicting intrinsically disordered regions

\# Reception and Impact

\#\# Scientific community

\#\# General public

\#\# Ethical considerations

\# Future Directions

\#\# Enhancements in accuracy

\#\# Integration with other biological tools

\#\# Expansion to other molecular complexes

\end{tcolorbox}
\caption{The complete outline of AlphaFold generated by \modelname.}
\label{case_outline}
\end{figure*}
\begin{figure*}

\begin{tcolorbox}[
colback=gray!10!white,
colframe=black!75!black,
colbacktitle=black!60!white, 
title=AlphaFold,  
width=\textwidth, 
]

    Background: Development
    
    Development: Algorithm,
    
    Algorithm: Protein Structure Predictions, Competitions and Benchmarks, Limitations
    
    Protein Structure Predictions: Applications, Limitations

    Competitions and Benchmarks: None
    
    Applications: Database,Reception and Impact
    
    Database: Source Code and Open Access
    
    Source Code and Open Access: None
    
    Limitations: Reception and Impact
    
    Reception and Impact: Future Directions
    
    Future Directions: None

\end{tcolorbox}
\caption{The writing plan for AlphaFold generated by \modelname. Each section is a dependency for the sections listed as its values; "None" indicates that the section does not serve as a dependency for any other section.}
\label{case_plan}
\end{figure*}




\begin{figure*}

\begin{tcolorbox}[
colback=gray!10!white,
colframe=black!75!black,
colbacktitle=black!60!white, 
title=AlphaFold,  
width=\textwidth, 
]
\# Development

\#\# Origin of the Project

The development of AlphaFold emerged from a long-standing challenge in the field of structural biology: accurately predicting protein structures based on amino acid sequences, a problem often referred to as the "protein-folding problem." This issue had persisted as one of the most significant puzzles in biology for over five decades, with limited progress despite numerous scientific efforts[2]. DeepMind, a subsidiary of Alphabet Inc., initiated the AlphaFold project with the goal of leveraging advances in artificial intelligence, particularly deep learning, to address this challenge. The project culminated in the release of AlphaFold 2 in 2020, a system that achieved near-experimental accuracy in many cases and fundamentally reshaped the landscape of protein structure prediction[3].\\

\#\# Collaboration with Scientific Communities

The development and adoption of AlphaFold have been significantly bolstered by its integration with and support from scientific communities. Following its launch, AlphaFold quickly became a focal point of discussion and collaboration within the structural biology field. For instance, researchers like Pedro Beltrao from ETH and EMBL-EBI initiated joint studies based on results shared on social media platforms, turning the excitement surrounding AlphaFold into a collective research effort. This bottom-up approach exemplified the enthusiasm and openness fostered by AlphaFold's innovations, allowing scientists from various institutions to collaborate and build upon shared findings[4].
Moreover, AlphaFold embraced the principles of open science by releasing its predictions and source code, alongside the establishment of an open-access protein structure database containing over 200 million protein structures. This democratized access to cutting-edge resources and enabled researchers globally to integrate AlphaFold's predictions into their work, regardless of institutional funding or geographic location. By adhering to transparency, accessibility, and collaboration, AlphaFold not only advanced the field of protein structure prediction but also inspired open science initiatives in other disciplines[2].\\

\# Algorithm

\#\# AlphaFold 1

The first iteration of AlphaFold debuted in the Critical Assessment of Structure Prediction (CASP13) competition, marking a significant milestone in protein structure prediction. While AlphaFold 1 relied on traditional homology modeling approaches combined with deep learning techniques, its ability to predict protein structures with reasonable accuracy laid the groundwork for future advancements. This version introduced key innovations in handling sequence alignments and template-based predictions, setting the stage for subsequent breakthroughs[5].\\

\#\# AlphaFold 2

AlphaFold 2 represented a transformative leap in protein structure prediction. Unveiled during CASP14, it utilized a completely redesigned architecture to achieve near-experimental accuracy in many cases. Unlike its predecessor, AlphaFold 2 did not rely solely on homology modeling or templates but could predict structures for previously unknown protein folds[6][5].
The model introduced innovations such as the incorporation of attention mechanisms and pairwise geometric features, enabling it to accurately model spatial relationships between amino acids. These advances allowed AlphaFold 2 to excel at decoupling the training and inference tasks, a unique approach that optimized the system for both learning from data and making predictions[7][6].

\end{tcolorbox}
\caption{A segment of the AlphaFold article generated by \modelname.}
\label{case_article}
\end{figure*}

\end{document}